\documentclass[a4paper,fleqn]{cas-sc}

\usepackage[authoryear]{natbib}

\usepackage{subcaption}
\usepackage{xcolor}
\usepackage{soul}
\PassOptionsToPackage{colorlinks=true}{hyperref}
\usepackage{cleveref}
\usepackage{makecell}
\usepackage{fancyvrb}
\usepackage{graphicx}
\usepackage{enumitem}
\usepackage{caption}   
\usepackage{xparse}    
\usepackage{algorithm,algpseudocode,float}
\usepackage{amsmath,amssymb}



\newcommand{\vect}[1]{\boldsymbol{#1}}

\hypersetup{
    colorlinks = true,    
    linkcolor = blue,     
    citecolor = blue,     
    urlcolor = blue,      
    filecolor = blue,     
    allcolors = blue,     
}

\crefname{figure}{\textcolor{blue}{Fig.}}{\textcolor{blue}{Fig.}} 
\crefname{table}{\textcolor{blue}{Table}}{\textcolor{blue}{Tables}}
\crefname{equation}{\textcolor{blue}{Eq.}}{\textcolor{blue}{Eqs.}}
\crefname{algorithm}{\textcolor{blue}{algorithm}}{\textcolor{blue}{algorithms}}
\Crefname{algorithm}{\textcolor{blue}{Algorithm}}{\textcolor{blue}{Algorithms}}

\def\tsc#1{\csdef{#1}{\textsc{\lowercase{#1}}\xspace}}
\tsc{WGM}
\tsc{QE}
\tsc{EP}
\tsc{PMS}
\tsc{BEC}
\tsc{DE}

\begin{document}

\let\WriteBookmarks\relax
\def\floatpagepagefraction{1}
\def\textpagefraction{.001}

\shorttitle{UniDrive for Interpretable Risk Understanding}

\shortauthors{Gao et~al.}

\title [mode = title]{UniDrive: A Unified Vision-Language and Grounding Framework for Interpretable Risk Understanding in Autonomous Driving}                      



%
\author[1,2]{Xiaowei Gao}




\affiliation[1]{organization={Department of Earth Science \& Engineering, Imperial College London},
    city={London},
    postcode={SW7 2AZ}, 
    country={United Kingdom}}

\affiliation[2]{organization={SpaceTimeLab, Department of Civil, Environmental, and Geomatic Engineering, University College London},
    city={London},
    postcode={WC1E 6BT}, 
    country={United Kingdom}}
    
\author[3]{Pengxiang Li}

\affiliation[3]{organization={Department of Computing, The Hong Kong Polytechnic University},
    city={Hong Kong},
    country={China}}
    
\author[2]{Yitai Cheng}

\author[4]{Ruihan Xu}

\affiliation[4]{organization={Trinity College, University of Oxford},
    city={Oxford},
    postcode={OX1 3BH}, 
    country={United Kingdom}}
\author%
[2]
{James Haworth}

\author%
[5]
{Stephen Law}

\affiliation[5]{organization={Department of Geography, University College London},
    city={London},
    postcode={WC1E 6BT}, 
    country={United Kingdom}}

\author%
[6,2]
{Yun Ye}[orcid=0000-0002-3346-4640]
\cormark[1]
\ead{yun.ye@ucl.ac.uk}
\affiliation[6]{organization={Centre for Global Infrastructure Resilience, The Bartlett School of Sustainable Construction, University College London},
   city={London},
   postcode={WC1E 7HB}, 
    country={United Kingdom}}
    
\cortext[cor1]{Corresponding author}



\begin{abstract}
Recent multimodal large language models (MLLMs) have shown strong potential for autonomous driving scene understanding, yet existing methods still face a fundamental trade-off between temporal reasoning and spatial precision. Models that rely on single-frame or low-resolution inputs often miss small, distant, or partially occluded hazards, while language-centric driving models frequently provide limited grounded evidence for their explanations. To address this gap, we propose UniDrive, a unified visual-language and grounding framework for interpretable risk understanding in autonomous driving. UniDrive combines a temporal reasoning branch that models scene dynamics from multi-frame visual input with a high-resolution perception branch that preserves fine-grained spatial details from the latest frame. The two branches are integrated through a gated cross-attention fusion module, enabling dynamic context to be aligned with precise spatial evidence. Based on the fused representation, UniDrive jointly generates natural-language risk descriptions and grounded bounding-box outputs for risk objects. Experiments on the DRAMA-Reasoning benchmark show that UniDrive outperforms representative image-based and video-based baselines in both captioning and risk-object grounding. In particular, UniDrive achieves the best overall performance on the validation split and demonstrates clear advantages in small-object localization, zero-shot generalization to NuScenes and BDD100K, and human-rated interpretability and trustworthiness. These results suggest that explicitly combining temporal semantics and high-resolution perception provides a stronger foundation for interpretable and safety-oriented autonomous driving systems. The code is available at https://github.com/pixeli99/unidrive-dev\href{https://github.com/pixeli99/unidrive-dev}.
\end{abstract}




\maketitle
\section{Introduction}

Autonomous driving systems are increasingly expected to operate in complex, dynamic, and safety-critical traffic environments \citep{che2026enhancing,ye2026wait,ye2026ehmis}. In such environments, safe driving requires more than detecting surrounding objects or producing low-level control commands. A reliable autonomous system should be able to understand which object or agent creates risk, how the risk evolves over time, where the risk-critical object is located, and why the situation requires a cautious response. This capability is particularly important in long-tail driving scenarios involving subtle hazards, distant objects, partial occlusions, vulnerable road users, uncertain intentions, and rapidly evolving interactions \citep{chen2024endtoend}. In these cases, a system that only outputs an opaque driving decision or a coarse scene description may be insufficient for safety validation, post-hoc auditing, and communication with human users. Therefore, interpretable and visually grounded risk understanding has become a critical capability for practical autonomous driving systems \citep{malla2023drama,zhou2024vision}.

Recent progress in vision-language models (VLMs) and multimodal large language models (MLLMs) provides a promising foundation for addressing this challenge. General-purpose multimodal models have demonstrated strong capabilities in visual reasoning, instruction following, visual grounding, and video understanding \citep{alayrac2022flamingo,liu2024improved,peng2023kosmos2,zhang2023videollama}. Motivated by these advances, driving-oriented MLLMs have extended language-enhanced perception to risk localization, scene captioning, visual question answering, high-resolution understanding, and end-to-end driving assistance \citep{malla2023drama,xu2024drivegpt4,ding2023hilmd,sima2024drivelm,tian2024drivevlm,hwang2024emma}. These studies suggest that language-enhanced visual understanding can improve the transparency, generalization, and human interpretability of autonomous driving systems.

Nevertheless, existing driving-oriented MLLM methods still face an important limitation for safety-critical risk understanding as they often struggle to jointly capture temporal risk evolution and fine-grained spatial evidence. Models based on single-frame reasoning or relatively low-resolution visual inputs can generate plausible descriptions of traffic scenes, but they may miss small, distant, or partially occluded risk objects that are critical for safe decision-making \citep{ding2023hilmd,xu2024drivegpt4}. Video-based models are better suited to capturing dynamic interactions and temporal context, yet their spatial grounding can remain coarse when risk objects occupy only a small region of the image \citep{zhang2023videollama}. Conversely, approaches that emphasize high-level language interaction, question answering, or driving interfaces may improve reasoning and communication, but they often provide limited support for fine-grained visual grounding of risk-critical targets \citep{sima2024drivelm,tian2024drivevlm,wang2023drivemlm}. As a result, current models may produce reasonable language explanations without precise visual evidence, or localize objects without adequately explaining their dynamic risk relevance.

This limitation is especially problematic because driving risk understanding is inherently both temporal and localized. A pedestrian stepping out from behind a parked vehicle, a cyclist gradually entering the ego lane, or a distant traffic participant with subtle motion cues requires the model to integrate temporal semantics with high-resolution spatial details. For safety-oriented interpretation, these two forms of information should not be treated as separate outputs. Instead, the model should align dynamic context with explicit visual evidence and generate explanations that are both semantically meaningful and spatially grounded. This requirement is closely aligned with risk-centric benchmarks such as DRAMA, where risk explanation and object-level grounding are both essential for understanding interactive driving scenarios \citep{malla2023drama}.

To address this gap, we propose \emph{UniDrive}, a unified visual-language and grounding framework for interpretable risk understanding in autonomous driving. UniDrive combines a temporal reasoning branch that models scene dynamics from multi-frame visual input with a high-resolution perception branch that preserves fine-grained spatial details from the latest frame. The two branches are integrated through a gated cross-attention fusion module, enabling the model to dynamically align temporal context with precise spatial evidence. Based on the fused representation, UniDrive jointly generates natural-language risk descriptions and grounded bounding-box outputs for risk objects. In this way, the model not only identifies hazardous agents, but also explains the underlying reason and links that explanation to explicit visual evidence.

We evaluate UniDrive on an extended risk-reasoning setting built upon DRAMA \citep{malla2023drama}. The original DRAMA dataset provides risk-object annotations for interactive driving scenarios, while our extended DRAMA-Reasoning setting enriches these annotations with textual descriptions of hazardous objects, ego-vehicle intentions, and safe driving suggestions. This setting enables joint evaluation of two safety-relevant capabilities: interpretable risk captioning and object-level visual grounding. Experiments show that UniDrive consistently improves both language-based risk explanation and grounded risk-object localization compared with representative image-based and video-based baselines. Additional evaluations on unseen driving datasets further examine the model's zero-shot generalization, robustness under challenging conditions, and human-perceived usefulness, accuracy, and trustworthiness.

The main contributions of this paper are summarized as follows:
\begin{itemize}
    \item We propose UniDrive, a unified visual-language and grounding framework for interpretable risk understanding in autonomous driving. Unlike models that only generate scene-level descriptions or ungrounded explanations, UniDrive jointly produces natural-language risk descriptions and bounding-box evidence for risk-critical objects.

    \item We introduce a dual-branch architecture that combines temporal reasoning from multi-frame visual inputs with high-resolution spatial perception from the latest frame. A gated cross-attention fusion module is designed to align dynamic scene context with fine-grained visual evidence, thereby improving both temporal risk interpretation and object-level grounding.

    \item We develop an extended DRAMA-Reasoning experimental setting by enriching risk-object annotations with explanatory descriptions, ego-vehicle intentions, and safe-action suggestions, enabling joint evaluation of risk explanation and visual grounding.

    \item We conduct comprehensive experiments on the DRAMA-Reasoning benchmark and unseen driving datasets, including comparisons with representative MLLM baselines, zero-shot generalization tests, ablation studies, robustness analysis, efficiency analysis, qualitative analysis, and human-centered evaluation. The results demonstrate that explicitly combining temporal semantics and high-resolution spatial evidence provides a stronger foundation for interpretable and safety-oriented autonomous driving.
\end{itemize}

The remainder of this paper is organized as follows. Section 2 reviews related work on multimodal reasoning and grounding, MLLMs for autonomous driving risk understanding, and benchmark-oriented safety evaluation. Section 3 presents the proposed UniDrive framework, including the temporal reasoning branch, the high-resolution perception branch, the spatio-temporal fusion module, and the unified reasoning and grounding mechanism. Section 4 reports the experimental setup and evaluation results, including comparisons with representative baselines, zero-shot generalization, ablation studies, human-centered evaluation, robustness analysis, efficiency analysis, and qualitative results. Section 5 discusses the main findings, underlying mechanisms, safety implications, limitations, and future research directions. Section 6 concludes the paper.

\section{Related Work}

\subsection{Multimodal Reasoning and Visual Grounding}

Recent advances in VLMs and MLLMs have substantially improved the ability of artificial intelligence systems to connect visual perception with language-based reasoning. Representative models such as Flamingo demonstrated that large language models can be conditioned on visual inputs through cross-attention, enabling few-shot visual-language understanding across diverse tasks \citep{alayrac2022flamingo}. LLaVA and its improved variants further showed that visual instruction tuning can produce strong multimodal conversational and reasoning abilities \citep{liu2024improved}. Beyond general visual dialogue, grounded multimodal generation has received increasing attention. Kosmos-2 links language outputs to visual regions for grounded generation \citep{peng2023kosmos2}, GroundHog strengthens the coupling between language generation and dense visual understanding \citep{zhang2024groundhog}, and SpatialRGPT highlights the importance of region-level geometry and spatial relationships for spatial reasoning \citep{cheng2024spatialrgpt}. Meanwhile, Video-LLaMA extends multimodal language modeling from static images to temporally structured video inputs, showing the potential of MLLMs for dynamic scene understanding \citep{zhang2023videollama}.

These studies provide important foundations for language-enhanced perception, visual grounding, and temporal reasoning. However, general-purpose MLLMs are not specifically designed for safety-critical autonomous driving scenarios. Driving risk understanding requires more than generic object recognition, image captioning, or visual dialogue. A model should identify which object creates risk, explain why it is hazardous, and ground this explanation in the visual scene. It should also account for temporal cues such as motion, interaction, and intention changes. Therefore, although existing MLLMs establish the feasibility of multimodal reasoning and grounding, their direct application to autonomous driving remains limited without mechanisms that explicitly integrate temporal risk evolution with fine-grained spatial evidence.

\subsection{MLLMs for Autonomous Driving Risk Understanding}

Driven by the progress of general-purpose VLMs and MLLMs, recent studies have adapted multimodal language models to autonomous driving. These works extend language-enhanced perception to tasks such as scene captioning, visual question answering, risk localization, planning-oriented reasoning, and end-to-end driving assistance. DRAMA introduced a benchmark for joint risk localization and captioning, emphasizing that risk understanding in driving scenes should couple natural-language explanation with object-level localization \citep{malla2023drama}. DriveGPT4 explored interpretable end-to-end autonomous driving by generating language outputs from driving videos \citep{xu2024drivegpt4}, while HiLM-D highlighted the value of high-resolution visual inputs for detailed driving scene understanding \citep{ding2023hilmd}. DriveLM formulated driving understanding as graph-based visual question answering, and DriveVLM investigated the broader convergence between autonomous driving and large vision-language models \citep{sima2024drivelm,tian2024drivevlm}. More recent systems, such as EMMA and BEV-injected multimodal driving models, further extend MLLMs toward end-to-end driving and bird's-eye-view enhanced understanding \citep{hwang2024emma,ding2024holistic}.

Another related direction uses large language models as language-centric agents for decision-oriented driving. A Language Agent for Autonomous Driving models driving as a cognitive process involving perception, memory, reasoning, and action selection \citep{mao2023languageagent}. LMDrive, Drive like a Human, and DriveMLM incorporate language or planning-oriented supervision into autonomous driving pipelines \citep{shao2024lmdrive,fu2024drivelikehuman,wang2023drivemlm}, while KOMA explores knowledge-driven multi-agent coordination through large language models \citep{jiang2024koma}. These methods are valuable for planning, control, and interaction, but their primary objective is often decision generation rather than risk-focused visual grounding. In safety-critical perception, an explanation is incomplete if it recommends an action without identifying the responsible risk object, its location, and the temporal context in which the risk emerges.

Despite these advances, existing driving-oriented MLLMs still leave an important gap for risk-centric autonomous driving. Many methods emphasize high-level semantic understanding, question answering, or planning interfaces, while the joint modeling of temporal hazard evolution and fine-grained grounded localization remains insufficiently explored. Video-based models can capture scene dynamics, but their spatial evidence may remain coarse when the risk object is small, distant, or partially occluded. High-resolution models improve visual detail, but they often provide limited reasoning about how a hazard emerges and evolves over time. In contrast, UniDrive treats driving risk understanding as a joint reasoning-grounding problem. It explicitly aligns temporal context with high-resolution spatial cues to generate both natural-language risk explanations and grounded bounding-box outputs.

\subsection{Benchmarks, Safety Evaluation, and Research Gap}

The development of driving-oriented benchmarks and safety evaluation frameworks has accelerated research on language-enhanced autonomous driving. Existing surveys have summarized the growing landscape of MLLMs and VLMs for autonomous driving, covering perception, planning, grounding, interaction, and embodied decision-making tasks \citep{cui2024survey,zhou2024vision,yang2023llm4drive,li2025adreview}. Benchmark-oriented studies have also begun to evaluate how language models support autonomous driving under structured task settings. LaMPilot provides an open benchmark for autonomous driving with language model programs \citep{ma2024lampilot}, while OmniTester investigates language-driven scenario testing and evaluation pipelines for autonomous vehicles \citep{lu2025omnitester}. From a safety perspective, DriveSOTIF explores the use of multimodal large models for safety-oriented perception under the SOTIF framework, highlighting the relevance of MLLMs to perception-related safety risks \citep{huang2025drivesotif}.

However, many existing evaluations still focus on scene-level question answering, planning responses, language-program execution, or general perception performance. These settings are valuable, but they do not fully assess whether a model can provide safety-relevant explanations that are visually verifiable. For autonomous driving, a plausible language response is not sufficient if it is not grounded in the correct risk object. A safety-oriented MLLM should explain why a specific object is risky, localize that object precisely, and remain reliable under challenging conditions such as occlusion, low visibility, small-object hazards, and distribution shifts.

This paper addresses this evaluation gap by formulating autonomous driving risk understanding as a joint reasoning-grounding problem. Building on DRAMA \citep{malla2023drama}, we develop an extended DRAMA-Reasoning setting that enriches risk-object annotations with explanatory descriptions, ego-vehicle intentions, and safe-action suggestions. This setting allows us to evaluate whether model outputs are not only linguistically plausible, but also grounded in the correct risk objects. Together with zero-shot transfer, robustness analysis, and human-centered evaluation, our experiments examine the reliability of UniDrive under safety-relevant conditions.

\section{Methods}

\begin{figure*}[t]
    \centering
    \includegraphics[width=1.0\textwidth]{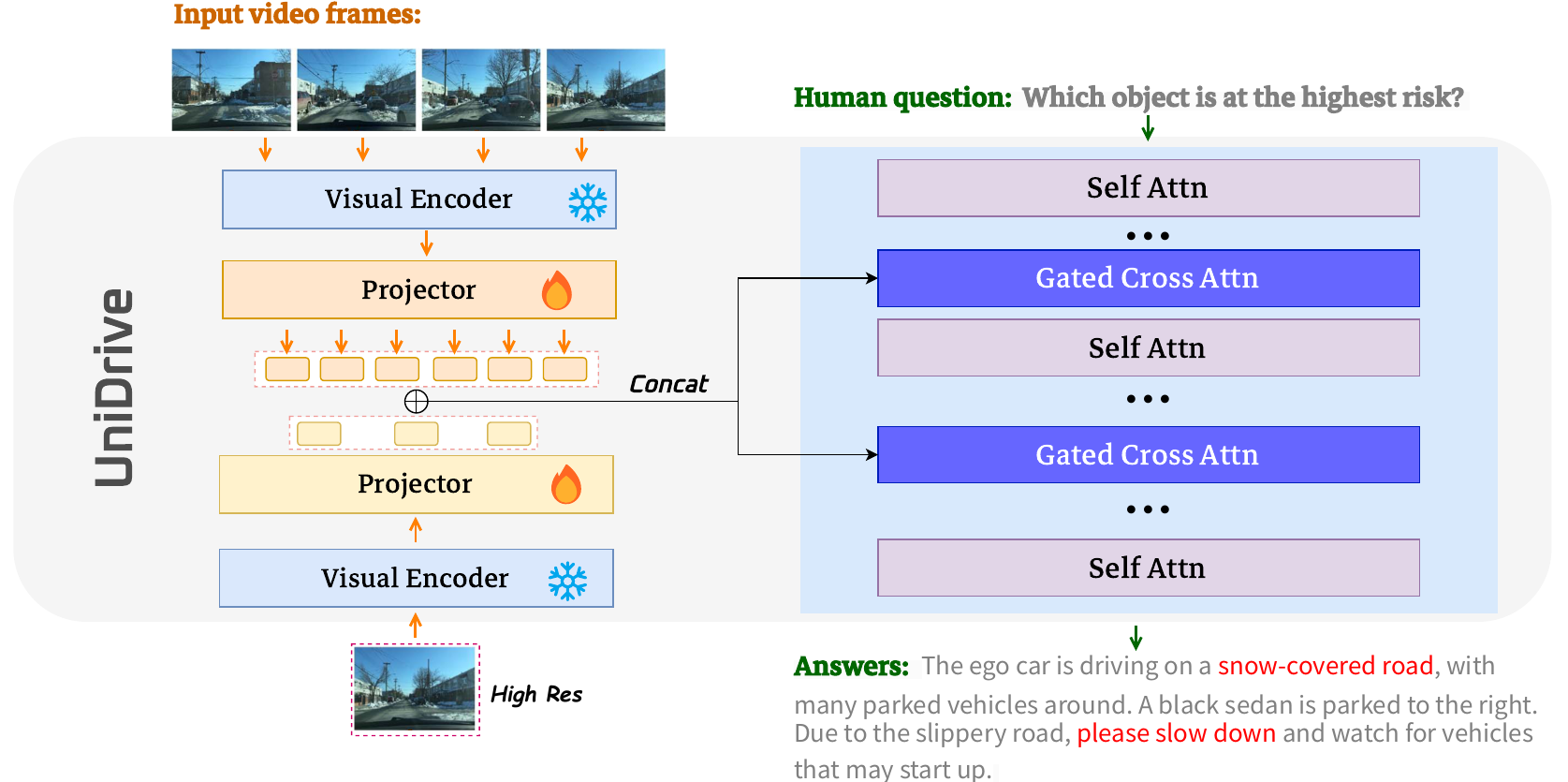}
    \caption{\textbf{The UniDrive Architecture.} Our model processes multi-view video inputs through two main pathways: a \textbf{Temporal Reasoning Branch (T-RB)} for semantic understanding of dynamics and a \textbf{High-Resolution Perception Branch (P-B)} for detailed spatial feature extraction from the most recent frame. A \textbf{Spatio-Temporal Fusion} module, using gated cross-attention, integrates these streams. This allows the Large Language Model (LLM) to generate both descriptive captions (identifying risks, causes, and ego-car actions) and precise, grounded bounding box predictions for risk objects in an end-to-end manner.}
    \label{fig:pipeline}
\end{figure*}
\subsection{Overview}
This section presents our UniDrive approach, a unified, end-to-end model designed to perceive, reason, and act in complex driving scenarios using multi-view video inputs. As illustrated in Figure~\ref{fig:pipeline}, our architecture is composed of two synergistic components: (a) a \textbf{Temporal Reasoning Branch (T-RB)} that analyzes video semantics to produce captions about risk objects, their causes, and the intended actions of the ego-car; and (b) a \textbf{High-Resolution Perception Branch (P-B)} that extracts fine-grained feature maps from high-resolution (HR) images to enhance perceptual accuracy, especially for small or distant objects. A core fusion module integrates these two information streams before they are processed by a large language model to generate the final output.

\subsection{Temporal Reasoning Branch (T-RB)}
The primary function of the T-RB is to process a sequence of low-resolution video frames to understand the temporal dynamics of a scene. It generates a high-level understanding of the situation, including identifying potential risks and inferring intentions. As depicted in Figure~\ref{fig:pipeline}, the T-RB consists of a vision encoder and a language reasoning model.
\subsubsection{Visual Encoder}
The visual encoder transforms raw video inputs into a sequence of visual tokens that the LLM can interpret. We build this encoder upon a CLIP-pretrained ViT-L/14, which remains frozen during training to preserve its strong generalization capabilities.

Formally, for a given video clip $V = \{v_1, v_2, \dots, v_L\}$ with $L$ frames, the ViT maps each frame $v_i$ to its $k$-th layer feature representation, $f_i^k$. This results in a set of features for the entire clip:
\begin{equation}
    F_k = \{f_1^k, f_2^k, \dots, f_L^k\}, \quad \text{where} \quad f_i^k \in \mathbb{R}^{N_f \times D_f}
\end{equation}
with $N_f$ being the number of patches (tokens) per frame and $D_f$ the feature dimension.

Before these features are passed to the LLM, a lightweight \textbf{feature aggregator} pools them along the temporal axis to create a compact representation:
\begin{equation}
    F^{\text{aggr}} = \text{Aggregator}(F_k)
\end{equation}
This stage serves two critical purposes: (1) it leverages temporal prior knowledge to create a holistic video representation, and (2) it significantly reduces the number of visual tokens, ensuring the total sequence length does not exceed the LLM’s context window limit.
\subsubsection{Language Reasoning Model (LRM)}
We adopt a Flamingo-style architecture, which employs alternating layers of Self-Attention and Gated Cross-Attention. This design is particularly adept at handling both textual and visual inputs. The Gated Cross-Attention layers allow the LLM to efficiently condition its text generation on the visual tokens $F^{\text{aggr}}$ provided by the encoder. 

Given a textual prompt $T$, the model generates a descriptive caption $Y = \{y_1, y_2, \dots, y_O\}$ by maximizing the conditional probability, modeled autoregressively:
\begin{equation}
    P(Y | T, F^{\text{aggr}}) = \prod_{i=1}^{O} P(y_i | y_{<i}, T, F^{\text{aggr}})
\end{equation}
By leveraging the extensive world knowledge of the pre-trained LLM, we prompt the model to generate captions that include risk identification, causal explanations, and actionable suggestions.

\subsection{High-Resolution Perception Branch (P-B)}
To overcome the limitations of low-resolution video inputs, which can obscure small but critical details (e.g., distant pedestrians or debris), we introduce a parallel perception branch. The P-B is designed to process a single high-resolution image $v_L^{\text{HR}}$, typically the most recent frame in the video sequence, to capture fine-grained spatial information. This branch utilizes a separate, dedicated vision encoder, $\Phi_{\text{P-B}}$, for which we also employ a Vision Transformer (ViT) architecture, chosen for its strong feature extraction capabilities on static, high-resolution images.
This encoder extracts a high-fidelity feature map $F^{\text{HR}}$:
\begin{equation}
    F^{\text{HR}} = \Phi_{\text{P-B}}(v_L^{\text{HR}}), \quad \text{where} \quad F^{\text{HR}} \in \mathbb{R}^{N_{\text{HR}} \times D_{\text{HR}}}
\end{equation}
Crucially, these high-resolution features are not directly consumed by the LLM's main body but are instead injected into the model via the spatio-temporal fusion module, providing precise spatial cues to ground the temporal reasoning.

\subsection{Spatio-Temporal Fusion}
The synergy between the T-RB and P-B is unlocked by our spatio-temporal fusion mechanism. Instead of naively concatenating features, we employ a more sophisticated strategy using Gated Cross-Attention layers. In this setup, the aggregated temporal features $F^{\text{aggr}}$ from the T-RB act as the \textit{queries} ($Q$), while the high-resolution spatial features $F^{\text{HR}}$ from the P-B serve as the \textit{keys} ($K$) and \textit{values} ($V$), after being passed through linear projections. Temporal features encode {what} dynamic risk cues to look for, so placing them as queries lets the model use motion and interaction context as a guide to selectively retrieve the most safety-relevant fine-grained details from the high-resolution spatial map. 

The core of the fusion is a standard cross-attention mechanism:
\begin{equation}
    \text{Attention}(Q, K, V) = \text{softmax}\left(\frac{QK^T}{\sqrt{d_k}}\right)V
\end{equation}
where $d_k$ is the dimension of the keys. This is integrated with a gating mechanism that dynamically balances the attended high-resolution features with the original temporal features. The final fused representation $F^{\text{fused}}$ is computed as:
\begin{equation}
    F^{\text{fused}} = \alpha \cdot \text{Attention}(Q, K, V) + (1-\alpha) \cdot Q
\end{equation}
where $\alpha$ is a learned, element-wise gating parameter. This allows the model to learn a dynamic alignment, using the temporal context (the query) to selectively attend to the most relevant high-resolution spatial details (the key-value pairs). The resulting representation is thereby enriched with both temporal context and precise spatial grounding.

\subsection{Unified Reasoning and Generation}
The fused spatio-temporal representation $F^{\text{fused}}$ serves as the rich, contextual input for our core reasoning module (i.e., a LLM). Following the Flamingo architecture~\citep{alayrac2022flamingo}, we keep the pre-trained LM blocks frozen and interleave newly initialized Gated Cross-Attention layers at every $k$-th transformer block.
This design is particularly adept at handling both textual and visual inputs. The Gated Cross-Attention layers allow the LLM to efficiently condition its text generation on the fused visual tokens $F^{\text{fused}}$.

Given a textual prompt $T$, the model generates a descriptive caption $Y = \{y_1, y_2, \dots, y_O\}$ by maximizing the conditional probability, modeled autoregressively:
\begin{equation}
    P(Y | T, F^{\text{fused}}) = \prod_{i=1}^{O} P(y_i | y_{<i}, T, F^{\text{fused}}) 
\end{equation}
By leveraging the extensive world knowledge of the pre-trained LLM (e.g., Llama2-7B), we prompt the model to generate captions that include risk identification, causal explanations, and actionable suggestions.

\begin{figure*}[h]
    \centering
    \includegraphics[width=1.0\textwidth]{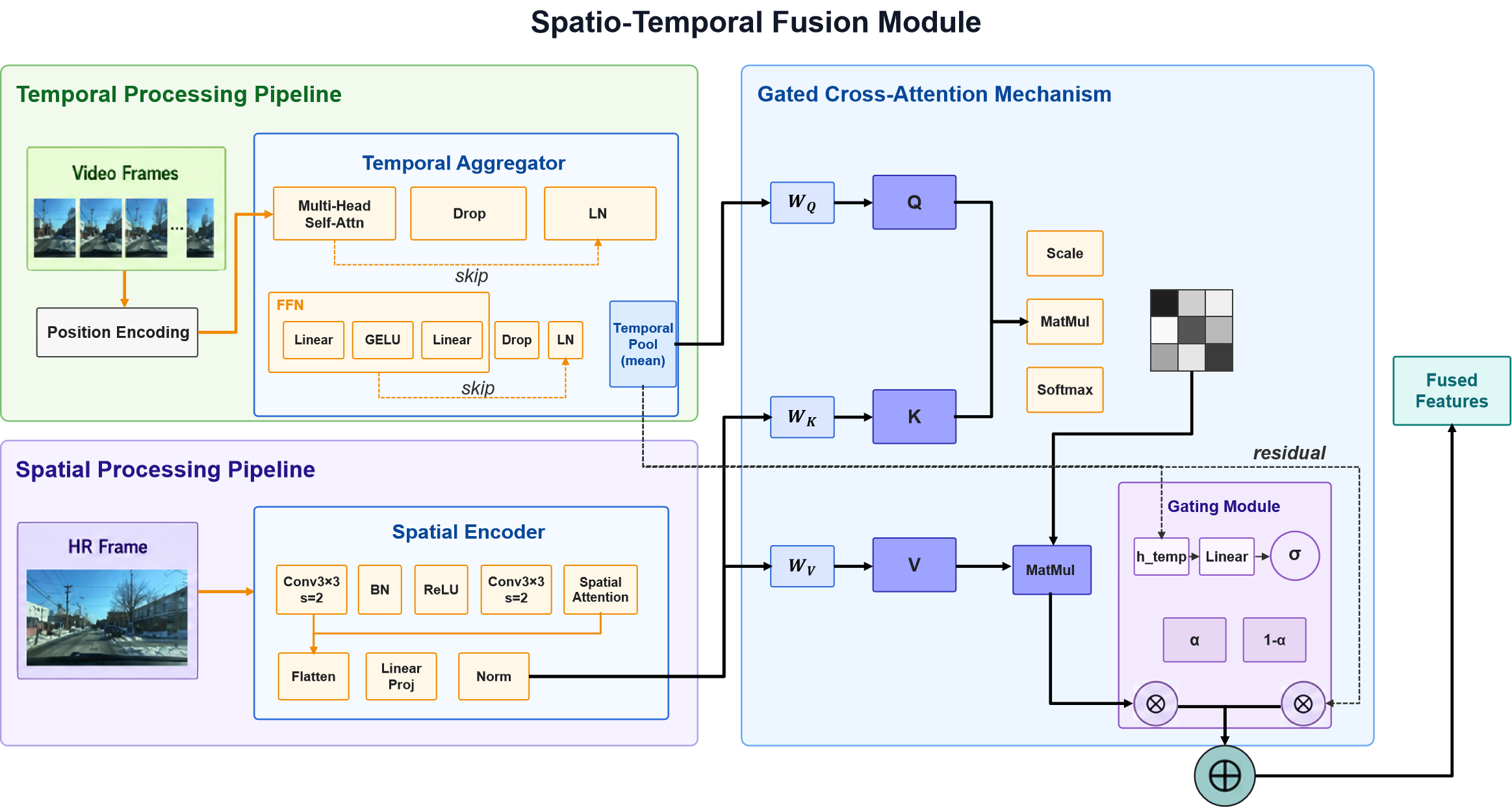}
    \caption{Detailed architecture of the Spatio-Temporal Fusion Module. Temporal features from multiple frames are aggregated and serve as queries (Q), while high-resolution spatial features are encoded to generate keys (K) and values (V) for the gated cross-attention mechanism. The gating parameter dynamically balances the contribution of attention-weighted features and residual temporal information.}
    \label{fig:stfusion}
\end{figure*}

\subsection{Grounding Output}
To seamlessly integrate object localization into the generative framework, we format bounding box coordinates as plain text. Specifically, we use special delimiters to represent a bounding box in the format \texttt{<box>x1, y1, x2, y2</box>}. This location information is directly appended to the end of the generated caption. This elegant approach avoids the need for auxiliary regression heads or separate detection decoders, allowing the entire model to be trained end-to-end with a standard language modeling objective and simplifying the overall architecture.

\subsection{Data and Training}
Our training process is straightforward, consisting of a single supervised fine-tuning (SFT) stage on a mixture of driving-related reasoning datasets. We start from the pre-trained weights of the vision encoders and the LLM, and we do not use reinforcement learning from human feedback (RLHF).

The primary training corpus is our newly curated \textbf{DRAMA-Reasoning} dataset. We build upon the original DRAMA dataset, which provides risk object annotations for 17,785 interactive driving scenarios. We extend these annotations by manually adding detailed textual descriptions of the ego-vehicle's intentions and recommended safe driving actions for each scenario. During the SFT stage, all model parameters are updated. To account for the different natures of the two branches, we set distinct initial learning rates, whereas $1\times 10^{-4}$ for the temporal reasoning branch to carefully fine-tune the pre-trained weights, and a higher learning rate of $4\times 10^{-4}$ for the perception branch components to allow them to adapt more quickly.



    
    
    
\begin{algorithm*}[t] 
\caption{UniDrive: Unified Forward Pass for Perception and Reasoning}
\label{alg:unidrive_formal_wide}
\begin{algorithmic}[1]
\Require 
    Video Clip $\mathcal{V} = \{v_t\}_{t=1}^L$, where $v_t \in \mathbb{R}^{H \times W \times 3}$; 
    High-resolution final frame $v_L^{\text{HR}} \in \mathbb{R}^{H' \times W' \times 3}$;
    Textual Prompt $\mathcal{P} = \{p_i\}_{i=1}^{M}$.
\Require 
    Temporal Encoder $\Phi_{\text{Temp}}: \mathbb{R}^{H \times W \times 3} \to \mathbb{R}^{N_p \times D_f}$;
    Spatial Encoder $\Phi_{\text{Spat}}: \mathbb{R}^{H' \times W' \times 3} \to \mathbb{R}^{N'_p \times D_h}$;
    Feature Aggregator $\mathcal{A}: \{\mathbb{R}^{N_p \times D_f}\}_{t=1}^L \to \mathbb{R}^{N_q \times D_f}$;
    Gated Cross-Attention Fusion Operator $\mathcal{F}_{\text{fuse}}$;
    Multimodal Large Language Model $\mathcal{M}_{\text{LLM}}$.
\Ensure 
    Generated textual output sequence $\mathcal{O}$.

\Statex
\Procedure{UniDrive\_Forward}($\mathcal{V}, v_L^{\text{HR}}, \mathcal{P}$)
    \Statex \Comment{\textit{Phase 1: Dual-Stream Asynchronous Feature Extraction}}
    
    \State \Comment{Temporal Reasoning Branch (T-RB) processing}
    \State Let the set of frame-wise feature maps be $\mathbf{F}_{\text{frames}} = \bigoplus_{t=1}^{L} \Phi_{\text{Temp}}(v_t)$.
    \State Derive the temporally aggregated query representation: $\vect{Q} \leftarrow \mathcal{A}(\mathbf{F}_{\text{frames}}) \in \mathbb{R}^{N_q \times D_f}$.
    
    \Statex
    \State \Comment{High-Resolution Perception Branch (P-B) processing}
    \State Extract fine-grained spatial features from the final high-resolution frame:
    $\mathbf{F}_{\text{HR}} \leftarrow \Phi_{\text{Spat}}(v_L^{\text{HR}}) \in \mathbb{R}^{N'_p \times D_h}$.
    
    \Statex \Comment{\textit{Phase 2: Spatio-Temporal Manifold Fusion via Gated Attention}}
    
    \State \Comment{Project features into Key-Value space for contextual grounding}
    \State Define projection matrices $W_K \in \mathbb{R}^{D_h \times d_k}, W_V \in \mathbb{R}^{D_h \times d_v}$.
    \State $\vect{K} \leftarrow \mathbf{F}_{\text{HR}} W_K$; $\vect{V} \leftarrow \mathbf{F}_{\text{HR}} W_V$.
    
    \State \Comment{Compute the attention-weighted high-resolution context}
    \State $\mathbf{A}_{\text{context}} \leftarrow \text{softmax}\left(\frac{\vect{Q}\vect{K}^\top}{\sqrt{d_k}}\right) \vect{V} \in \mathbb{R}^{N_q \times d_v}$.
    
    \State \Comment{Dynamically blend temporal and spatial information via a gating mechanism}
    \State Let $\sigma$ be the sigmoid function and $W_\alpha$ be learnable gating weights.
    \State $\vect{\alpha} \leftarrow \sigma(W_\alpha [\vect{Q}; \mathbf{A}_{\text{context}}])$, where $[\cdot;\cdot]$ denotes concatenation.
    \State The final fused representation is computed as a convex combination:
    $\mathbf{F}_{\text{fused}} \leftarrow \vect{\alpha} \odot \mathbf{A}_{\text{context}} + (1 - \vect{\alpha}) \odot (\vect{Q}W_Q)$, where $W_Q$ projects $\vect{Q}$ to dimension $d_v$.
    
    \Statex \Comment{\textit{Phase 3: Autoregressive Generation Conditioned on Fused Multimodal Context}}
    
    \State Initialize output sequence $\mathcal{Y}_0 = \emptyset$ and initial hidden state $h_0$.
    \State Tokenize and embed prompt: $\vect{E}_{\mathcal{P}} = \text{Embed}(\mathcal{P})$.
    
    \For{$k=1, \dots, K_{\text{max}}$}
        \State \Comment{Update model state and predict next token distribution}
        \State $P(\mathcal{Y}_k | \mathcal{Y}_{<k}, \mathbf{F}_{\text{fused}}, \vect{E}_{\mathcal{P}}) = \mathcal{M}_{\text{LLM}}(h_{k-1}, [\mathbf{F}_{\text{fused}}; \text{Embed}(\mathcal{Y}_{k-1})], \vect{E}_{\mathcal{P}})$.
        
        \State \Comment{Select the next token via a decoding strategy (e.g., nucleus sampling)}
        \State $y_k \sim P(\mathcal{Y}_k)$.
        
        \If{$y_k = \langle\text{eos}\rangle$}
            \State \textbf{break}
        \EndIf
        
        \State $\mathcal{Y}_k \leftarrow \mathcal{Y}_{k-1} \oplus y_k$. \Comment{$\oplus$ denotes sequence concatenation}
    \EndFor
    
    \State $\mathcal{O} \leftarrow \text{Detokenize}(\mathcal{Y}_K)$.
    \State \textbf{return} $\mathcal{O}$.
\EndProcedure
\end{algorithmic}
\end{algorithm*}

\section{Experiments and Results}
\label{sec:experiments}

In this section, we conduct a comprehensive evaluation of \emph{UniDrive}. Our experiments are designed to achieve three primary goals: 
1) To benchmark UniDrive against state-of-the-art multimodal models on challenging driving-related tasks.
2) To perform a rigorous ablation study to dissect the contribution of each core component of our architecture. 
3) To provide qualitative visualizations that offer intuitive insights into the model's reasoning and perceptual capabilities.
We evaluate \emph{UniDrive} on our newly curated \textsc{DRAMA-Reasoning} benchmark, covering both captioning and risk-object detection. Unless otherwise noted, all results are reported on the \textit{val} split, and all hyper-parameters follow the defaults described below.

\subsection{Experimental Setup}

\paragraph{Datasets.}
We conduct our primary experiments using the \textbf{DRAMA} dataset \citep{malla2023drama}, which contains 17,785 two-second-long interactive driving scenarios captured at 30 FPS. While DRAMA provides crucial bounding box annotations for risk-tagged objects, it lacks the rich, explanatory language needed to train and evaluate high-level reasoning. To address this, we have extended its annotations to create the \textbf{DRAMA-Reasoning} dataset. For each scenario, we manually added detailed captions under three classes, namely (1) the primary risk object and the reason it is hazardous, (2) the inferred intention of the ego-vehicle (e.g., "slowing down," "maintaining speed"), and (3) a safe driving suggestion (e.g., "prepare to brake"). This extension transforms the dataset into a comprehensive benchmark for both perception and interpretable reasoning. Further details on our annotation process are provided in the supplementary material.

\paragraph{Evaluation Metrics.}
The tasks on the DRAMA-Reasoning dataset include two main objectives: (1) a captioning task aimed at identifying and explaining risk objects while predicting the ego-vehicle's intended actions, and (2) a risk object detection task. We assess captioning performance using standard language generation metrics: \textbf{BLEU-4} (B4), \textbf{METEOR} (M), \textbf{CIDEr} (C), and \textbf{SPICE} (S) \cite{}. For detection performance, we measure the mean Intersection over Union (\textbf{mIoU}). To better understand performance across different object scales, we further categorize the IoU based on object area: small (\textbf{IoU$_\text{S}$}), medium (\textbf{IoU$_\text{M}$}), and large (\textbf{IoU$_\text{L}$}).

\paragraph{Implementation Details.}
We use Llama2-7B as the backbone LLM, initialized from its publicly available pre-trained checkpoint. Our method is implemented in PyTorch and trained on a system with 4 NVIDIA A100 80GB GPUs. For video inputs, we uniformly sample $L=5$ frames from each clip, ensuring the final frame is always included for bounding box prediction. Input frames are resized and center-cropped to a fixed dimension of $224 \times 224$ for the temporal reasoning branch and processed at a higher resolution for the perception branch. We employ the AdamW optimizer with a cosine annealing learning rate schedule. Reflecting the different initialization states of our model's components, we set initial learning rates of $1\times 10^{-4}$ for the pre-trained components of the reasoning branch and $4\times 10^{-4}$ for the newly added components in the high-resolution perception branch. The global batch size is set to 32.

\subsection{Comparison with State-of-the-Art Methods}
We compare UniDrive against a suite of leading image-based and video-based MLLMs. As shown in Table~\ref{tab:main}, UniDrive establishes a new state of the art across both captioning and detection tasks.

\paragraph{Comparison with Image-Based Models.}
First, we evaluate against models that process only a single image frame. Even when our model is constrained to a single-frame input (denoted as "Ours w/o ST"), it already demonstrates a significant advantage. It surpasses strong baselines like InstructBLIP \citep{dai2023instructblip} and Shikra* \citep{chen2023shikra} by a large margin. For instance, our image-only model achieves an mIoU of \textbf{59.8}, a +9.5 improvement over the next best competitor, Shikra*. This highlights the efficacy of our high-resolution perception branch and the architectural design that effectively grounds linguistic reasoning in fine-grained visual details. The captioning scores are also state-of-the-art, with a CIDEr score of \textbf{246.7}, indicating that better spatial understanding directly translates to more accurate and relevant descriptions.

\paragraph{Comparison with Video-Based Models.}
When leveraging the full temporal context of video clips, UniDrive's performance advances further, underscoring the importance of temporal reasoning. Our full model significantly outperforms specialized video-language models like eP-ALM \citep{shukor2023ep} and Video-LLAMA \citep{zhang2023videollama}. UniDrive achieves a CIDEr score of \textbf{277.5} and a B4 score of \textbf{60.3}, showcasing its superior ability to generate coherent and contextually appropriate narratives about dynamic driving events. Most critically, the introduction of temporal information boosts detection performance to an mIoU of \textbf{61.2}, with a notable improvement in detecting small objects (IoU$_\text{S}$ of \textbf{31.0}). This suggests that temporal cues help the model anticipate and disambiguate transient road hazards that are difficult to identify from a single snapshot. The overall AVG score of \textbf{60.8} places UniDrive significantly ahead of all other methods, confirming the powerful synergy of its spatial and temporal processing capabilities.

\begin{table*}[ht]
\centering
\caption{\textbf{Comparison with the state-of-the-art on the DRAMA-Reasoning \textit{val} split.} UniDrive is benchmarked against leading image-based and video-based models. For all metrics, higher is better. ‘AVG’ is the arithmetic mean of B4 and mIoU, providing a balanced view of language and perception performance. Our model sets a new state of the art in all categories.}
\label{tab:main}
\begin{tabular}{c|c|cccc|cccc|c}
\toprule
\multirow{2}{*}{Input} & \multirow{2}{*}{Method} & \multicolumn{4}{c|}{Captioning} & \multicolumn{4}{c|}{Detection} & \multirow{2}{*}{AVG} \\
 &  & B4 & M & C & S & mIoU & mIoU$_S$ & IoU$_M$ & IoU$_L$ &  \\
\midrule
\multirow{5}{*}{Image} & BLIP-2 & 46.1 & 34.3 & 194.7 & 50.7 & 46.3 & 8.1 & 60.2 & 73.7 & 46.2 \\
 & LLaVA & 47.5 & 35.2 & 198.6 & 48.3 & 47.2 & 8.0 & 62.1 & 74.2 & 47.4 \\
 & InstructBLIP & 49.9 & 37.9 & 205.0 & 50.9 & 47.8 & 9.1 & 62.2 & 74.5 & 48.9 \\
 & Shikra* & 49.8 & 37.7 & 204.7 & 50.7 & 50.3 & 10.4 & 59.5 & 73.8 & 50.1 \\
 & Ours w/o ST & \textbf{55.2} & \textbf{38.1} & \textbf{246.7} & \textbf{54.3} & \textbf{59.8} & \textbf{29.8} & \textbf{64.3} & \textbf{82.1} & \textbf{57.5} \\
\midrule
\multirow{3}{*}{Video} & eP-ALM & 51.4 & 38.0 & 225.1 & 52.8 & 43.2 & 7.2 & 56.8 & 68.8 & 47.3 \\
 & Video-LLAMA & 53.9 & 37.8 & 229.5 & 52.6 & 42.8 & 6.9 & 55.3 & 67.9 & 48.4 \\
 & \textbf{UniDrive (Ours)} & \textbf{60.3} & \textbf{39.6} & \textbf{277.5} & \textbf{58.1} & \textbf{61.2}& \textbf{31.0} & \textbf{66.5} & \textbf{83.7} & \textbf{60.8} \\
\bottomrule
\end{tabular}
\end{table*}

\subsection{Generalization to Unseen Datasets and Scenarios}
A fundamental measure of a model's utility is its ability to generalize beyond its training distribution. To rigorously test this, we evaluate UniDrive's zero-shot performance on the \textbf{NuScenes} dataset \citep{caesar2020nuscenes}, a large-scale, industry-standard benchmark known for its diverse geographic locations, weather conditions, and complex traffic scenarios. This evaluation is critical as it simulates a real-world deployment scenario where the model must confront entirely new visual and contextual distributions without any prior fine-tuning.

We designed a challenging Visual Question Answering (VQA) task by curating a diverse set of questions from the NuScenes validation split. These questions were crafted to probe three distinct cognitive abilities: (1) \textbf{Object Presence}, which tests the model's core perceptual ability to identify specific objects (e.g., "Is there a stroller on the sidewalk?"); (2) \textbf{Traffic State}, which assesses its capacity to recognize critical semantic states (e.g., "What color is the traffic light for the ego-vehicle?"); and (3) \textbf{Situation Reasoning}, the most complex category, which demands an understanding of latent interactions, intentions, and potential risks (e.g., "Why is the car ahead of us braking?").

The results, presented in Table~\ref{tab:nuscenes_vqa}, demonstrate UniDrive's generalization capabilities. Compared to the strong Video-LLAMA baseline, UniDrive achieves a significantly higher average accuracy of \textbf{75.3\%}, marking a \textbf{7.2\%} improvement. The performance gains are consistent across all categories. The advantage in "Object Presence" (+5.6\%) can be attributed to our high-resolution perception branch, which is more adept at identifying objects in novel contexts. However, the most telling result is the substantial \textbf{+9.1\%} lead in "Situation Reasoning". This highlights the effectiveness of our spatio-temporal fusion mechanism and the temporal reasoning branch. These components enable UniDrive to not just see objects, but to understand their dynamic relationships and infer causality, a crucial skill for safe navigation. 
These zero-shot results suggest that UniDrive captures transferable cues for driving scene understanding rather than dataset-specific patterns, which is encouraging for deployment under distribution shift.

\begin{table}[h]
\centering
\caption{\textbf{Zero-shot VQA performance on the NuScenes dataset.} We report accuracy (\%) on different question types.The results show that UniDrive generalizes significantly better to unseen data, especially on complex reasoning tasks, without any fine-tuning.}
\label{tab:nuscenes_vqa}
\resizebox{\columnwidth}{!}{%
\begin{tabular}{l|ccc|c}
\toprule
\multirow{2}{*}{Method} & \multicolumn{3}{c|}{Question Type Accuracy (\%)} & \multirow{2}{*}{Average Acc. (\%)} \\
 & Object Presence & Traffic State & Situation Reasoning & \\
\midrule
Video-LLAMA & 71.3 & 78.5 & 54.5 & 68.1 \\
\textbf{UniDrive (Ours)} & \textbf{76.9} (+5.6) & \textbf{82.3} (+3.8) & \textbf{63.6} (+9.1) & \textbf{75.3} (+7.2) \\
\bottomrule
\end{tabular}%
}
\end{table}

\paragraph{Generalization to Diverse Driving Benchmarks.}
To further probe the model's generalization in a zero-shot setting, we also evaluated UniDrive on the challenging \textbf{BDD100K} dataset \citep{yu2020bdd100k}. We prompted the model to identify and localize the most significant risk in video clips from the validation set. As shown in Table~\ref{tab:bdd100k_results}, UniDrive demonstrates a strong ability to transfer its capabilities to this entirely new data distribution. It significantly outperforms the Video-LLAMA baseline in localizing critical road actors, especially vulnerable road users like pedestrians. Qualitatively, its generated risk descriptions were more context-aware and aligned with human driver intuition. This strong performance underscores that UniDrive's architecture learns generalizable principles of risk perception rather than memorizing dataset-specific patterns.

\begin{table}[h!]
\centering
\caption{\textbf{Zero-shot risk object detection on the BDD100K validation set.} We report mean Average Precision (mAP) at an IoU threshold of 0.5. UniDrive shows superior generalization to a new, diverse dataset.}
\label{tab:bdd100k_results}
\begin{tabular}{l|cccc}
\toprule
Method & mAP@.50 & AP\textsubscript{pedestrian} & AP\textsubscript{cyclist} & AP\textsubscript{car} \\ \midrule
Video-LLAMA & 41.3 & 35.8 & 30.1 & 58.0 \\
\textbf{UniDrive (Ours)} & \textbf{52.7} & \textbf{45.2} & \textbf{41.5} & \textbf{71.4} \\ \bottomrule
\end{tabular}
\end{table}

\subsection{Ablation Study}
\label{ssec:ablation}

To validate our design choices and understand the contribution of each technical ingredient in UniDrive, we perform a series of controlled ablations on the DRAMA-Reasoning \emph{val} split. As shown in Table~\ref{tab:ablation}, every component proves to be integral to the model's final performance.

\paragraph{Temporal Reasoning Branch (T-RB).}
Removing the T-RB and relying solely on the most recent frame forces the model to act as a pure image-based system. While the high-resolution branch maintains strong detection performance (59.8 mIoU), the captioning quality drops significantly (e.g., \textbf{-30.8} in CIDEr). This demonstrates that temporal context is vital for higher-level reasoning, such as anticipating future ego-car actions and explaining the evolution of a hazardous situation.

\paragraph{High-Resolution Perception Branch (P-B).}
Discarding the perception branch has the most dramatic impact on detection, with mIoU plummeting by \textbf{13.3} points. The effect is especially pronounced for small objects (IoU$_\text{S}$ drops from 31.0 to 12.4), confirming that fine-grained visual details from the HR stream are indispensable for precise spatial localization. Captioning also suffers, as the model struggles to ground its textual descriptions in accurate visual evidence, reaffirming the need for high-fidelity perception.

\paragraph{Spatio-Temporal Fusion (STF).}
We replace our proposed gated cross-attention fusion with a naive late concatenation of features from the two branches ("w/o STF"). This change leads to a substantial drop across all metrics, with mIoU falling to 44.6. This result strongly suggests that simply presenting both temporal and spatial features to the LLM is insufficient. The explicit, learned alignment facilitated by our fusion mechanism is critical for effectively integrating temporal semantics with high-resolution spatial cues.

\paragraph{Box Token Grounding.}
To verify the effectiveness of our lightweight \texttt{<box>} token, we ablate it and instead rely on naive string matching to find a predicted class name in the ground truth. As expected, all detection metrics collapse to zero, while captioning quality remains almost unchanged. This confirms that our direct-to-text bounding box representation provides an efficient and fully differentiable method for injecting localization supervision without interfering with the model's language generation capabilities.

Overall, these ablations reveal that each component contributes meaningfully to UniDrive's performance. In particular, the P-B and STF are key to accurate spatial grounding, whereas the T-RB is essential for generating temporally coherent and insightful descriptions.

\begin{table*}[ht]
\centering
\caption{\textbf{Ablation results on the DRAMA-Reasoning \textit{val} split.} Each row represents the removal of exactly one component from the full UniDrive model. AVG is the arithmetic mean of B4 and mIoU. The results demonstrate that all components are critical for achieving optimal performance.}
\label{tab:ablation}
\setlength{\tabcolsep}{5pt}
\begin{tabular}{l|cccc|cccc|c}
\toprule
\multirow{2}{*}{Model Variant} & \multicolumn{4}{c|}{Captioning$\uparrow$} & \multicolumn{4}{c|}{Detection$\uparrow$} & \multirow{2}{*}{AVG$\uparrow$} \\
 & B4 & M & C & S & mIoU & mIoU$_{\!S}$ & mIoU$_{\!M}$ & mIoU$_{\!L}$ &  \\
\midrule
Full \textbf{UniDrive}                & \textbf{60.3} & \textbf{39.6} & \textbf{277.5} & \textbf{58.1} & \textbf{61.2} & \textbf{31.0} & \textbf{66.5} & \textbf{83.7} & \textbf{60.8}\\
\quad w/o Temporal Reasoning Branch   & 55.2 & 38.1 & 246.7 & 54.3 & 59.8 & 29.8 & 64.3 & 82.1 & 57.5\\
\quad w/o High-Res Perception Branch  & 52.8 & 37.6 & 238.2 & 53.7 & 47.9 & 12.4 & 60.7 & 78.5 & 50.4\\
\quad w/o Spatio-Temporal Fusion      & 52.4 & 36.9 & 222.8 & 51.7 & 44.6 & 9.7 & 58.1 & 70.2 & 48.5\\
\quad w/o Box Token Grounding         & 60.1 & 39.5 & 276.1 & 57.9 &  \phantom{0}0.0 & \phantom{0}0.0 & \phantom{0}0.0 & \phantom{0}0.0 & 30.1\\
\bottomrule
\end{tabular}
\end{table*}

\subsection{Human-Centric Evaluation of Interpretability}
While automatic metrics measure objective accuracy, they do not capture the subjective quality and usefulness of the generated explanations for human users. To quantitatively evaluate UniDrive on the core claims of interpretability and trustworthiness, we conducted a formal user study.

\paragraph{Study Design.} 
We recruited 25 participants (aged 20-32, 12 male and 13 female; all with valid driving licenses, normal or corrected-to-normal vision, and at least two years of driving experience) through campus advertisements. We showed them 30 challenging scenarios from our validation set. For each scenario, we presented the explanations from UniDrive and Video-LLAMA in a randomized, side-by-side format and asked participants to choose the superior one based on three criteria: \textbf{Usefulness} (provides more valuable information for a safe decision), \textbf{Accuracy} (more precisely describes the scene), and \textbf{Trustworthiness} (which AI they would trust more in their own car).

\paragraph{Results.}
The results, summarized in Table~\ref{tab:user_study}, show a clear and strong human preference for UniDrive's explanations. Across all criteria, participants favored UniDrive over 74\% of the time, with the highest margin in Trustworthiness (82.0\%). This provides compelling evidence that the richer, more detailed, and temporally-aware explanations generated by our model are not only quantitatively better but are also perceived by humans as significantly more reliable and useful. This is a crucial step towards building human-AI trust in safety-critical autonomous systems.

\begin{table}[h!]
\centering
\caption{\textbf{Results of the human-centric evaluation study.} Scores show the percentage of times a model's explanation was chosen as superior by human evaluators.}
\label{tab:user_study}
\begin{tabular}{l|ccc}
\toprule
Criterion & UniDrive Win \% & Baseline Win \% & Tie / No Pref. \% \\ \midrule
Usefulness & \textbf{74.7\%} & 20.0\% & 5.3\% \\
Accuracy & \textbf{76.8\%} & 18.8\% & 4.4\% \\
Trustworthiness & \textbf{82.0\%} & 12.9\% & 5.1\% \\ \bottomrule
\end{tabular}

\end{table}
\subsection{Robustness, Efficiency, and Scalability}
Beyond accuracy on standard benchmarks, the practical viability of an autonomous driving model hinges on its robustness to adverse conditions and its computational efficiency. In this section, we analyze these critical aspects of UniDrive.

\paragraph{Robustness to Adverse Conditions.}
Driving environments are often non-ideal, with challenges like poor illumination and inclement weather that can severely degrade sensor input and cripple perception systems. To quantify UniDrive's resilience, we partitioned the DRAMA-Reasoning validation set into challenging subsets based on their metadata: \texttt{Day} vs. \texttt{Night} and \texttt{Clear} vs. \texttt{Rainy} weather. The results are detailed in Table~\ref{tab:robustness}. As expected, all models exhibit some performance degradation under these challenging conditions. However, UniDrive demonstrates a significantly more graceful degradation compared to the baseline. For example, in nighttime scenarios, UniDrive's mIoU drops by only 5.1 points (an 8\% relative decrease), whereas Video-LLAMA's performance falls by 8.6 points (a 20\% relative decrease). A similar trend is observed in rainy conditions. This enhanced robustness stems from our dual-branch design. The temporal reasoning branch can infer the presence and motion of objects even when they are partially obscured or have low contrast in a single frame, while the high-resolution branch maximizes the extraction of any available visual detail. This synergy makes UniDrive less susceptible to common visual corruptions, a vital feature for ensuring safety-critical reliability.

\begin{table}[h!]
\centering
\caption{\textbf{Robustness analysis on challenging subsets of DRAMA-Reasoning.} We report key captioning (CIDEr) and detection (mIoU) metrics. UniDrive shows superior resilience and more graceful performance degradation in adverse conditions.}
\label{tab:robustness}
\begin{tabular}{l|cc|cc}
\toprule
\multirow{2}{*}{Condition} & \multicolumn{2}{c|}{UniDrive (Ours)} & \multicolumn{2}{c}{Video-LLAMA (Baseline)} \\
& mIoU$\uparrow$ & CIDEr$\uparrow$ & mIoU$\uparrow$ & CIDEr$\uparrow$ \\
\midrule
Day / Clear (Base) & 61.2 & 277.5 & 42.8 & 229.5 \\
Night & 56.1 & 259.3 & 34.2 & 201.7 \\
Rainy & 57.5 & 265.8 & 36.1 & 210.4 \\
\bottomrule
\end{tabular}
\end{table}

\paragraph{Efficiency and Scalability.}
While performance is paramount, any model intended for eventual deployment must operate within reasonable computational budgets. In Table~\ref{tab:efficiency}, we analyze UniDrive's computational footprint. Although UniDrive has a moderately larger parameter count and higher GFLOPs due to its dual-branch architecture, it is designed with scalability in mind. By keeping the core LLM frozen and employing an efficient, lightweight fusion mechanism, the additional overhead is well-managed. Its inference speed of 9.8 FPS on a single NVIDIA A100 GPU is competitive and demonstrates a favorable trade-off between its significantly advanced reasoning capabilities and its computational cost. Furthermore, its Flamingo-style cross-attention architecture is inherently more scalable for handling even longer video sequences or a greater number of camera views compared to models relying solely on self-attention, which suffer from quadratic complexity. This suggests that UniDrive scales more favorably to longer sequences or additional camera views than self-attention-only models.

\begin{table}[h!]
\centering
\caption{\textbf{Efficiency and resource comparison.} We compare UniDrive with key baselines on model size, computational load, and inference speed. UniDrive achieves a strong balance between performance and practical efficiency.}
\label{tab:efficiency}
\begin{tabular}{l|ccc}
\toprule
Method & Parameters (B) & GFLOPs & Speed (FPS) \\
\midrule
LLaVA-1.5 (7B) & 7.1 & 795 & 12.1 \\
Video-LLAMA (7B) & 7.8 & 910 & 10.5 \\
\textbf{UniDrive (Ours)} & 8.2 & 980 & 9.8 \\
\bottomrule
\end{tabular}
\end{table}

\subsection{Qualitative Analysis}
\begin{figure*}[t]
    \centering
    \includegraphics[width=1.0\textwidth]{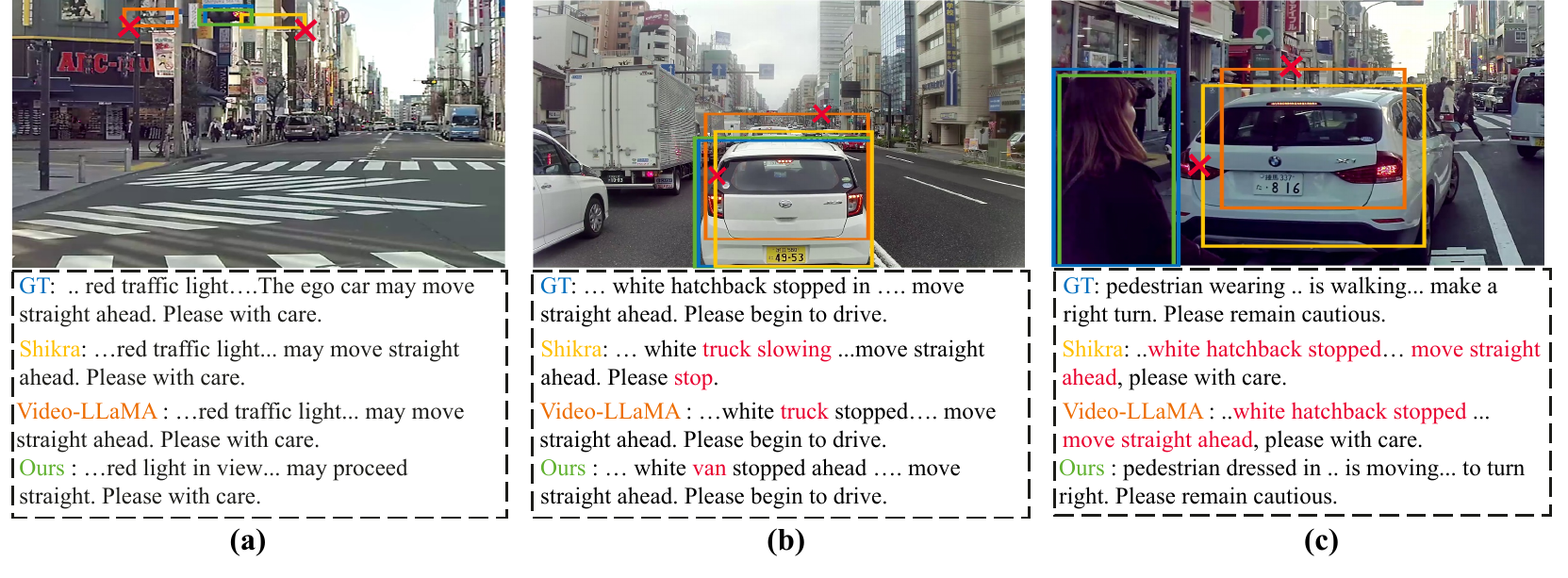}
    \caption{\textbf{Qualitative Comparison of UniDrive against a Video-LLM baseline (Video-LLAMA).} We showcase three challenging scenarios from the DRAMA-Reasoning val split. The outputs include the generated caption and the predicted bounding box (in green). (a) All models correctly identify the red traffic light, but UniDrive produces a more concise and action-oriented description. (b) Shikra incorrectly identifies the vehicle type and predicts the wrong ego-action ("stop"), while UniDrive correctly identifies the stopped white van and recommends the appropriate response. (c) UniDrive uniquely identifies the pedestrian as the primary risk object and correctly anticipates the right-turn maneuver, whereas both baselines focus on an irrelevant stopped vehicle. These examples highlight our model's enhanced capabilities in both fine-grained object recognition and safety-critical reasoning.}
    \label{fig:compare}
\end{figure*}

To complement our quantitative results, Figure~\ref{fig:compare} presents 
a qualitative comparison of UniDrive against Shikra and Video-LLaMA on 
three representative scenarios from the DRAMA-Reasoning \textit{val} split.

In scenario \textbf{(a)}, all models correctly identify the red traffic 
light as the key contextual cue. However, UniDrive produces a more concise 
and action-oriented description, directly linking the traffic state to the 
appropriate ego-vehicle response. In scenario \textbf{(b)}, the primary risk object is a white hatchback stopped ahead. Shikra misidentifies the vehicle type and incorrectly 
predicts the ego-action as ``stop'', while Video-LLaMA identifies the 
stopped vehicle but also misclassifies it as a truck. UniDrive correctly 
identifies the stopped white van and recommends the appropriate driving 
response of beginning to drive. In scenario \textbf{(c)}, the key hazard is a pedestrian crossing the road. Both Shikra and Video-LLaMA fail to identify the pedestrian, instead 
focusing on an irrelevant stopped vehicle. In contrast, UniDrive uniquely 
identifies the pedestrian as the primary risk object, correctly describes 
their motion, and anticipates the right-turn maneuver required by the 
ego-vehicle. These examples intuitively illustrate how UniDrive's 
architectural innovations translate into more reliable and safety-oriented 
driving intelligence.

\begin{figure*}[t]
  \centering
  \includegraphics[width=\textwidth]{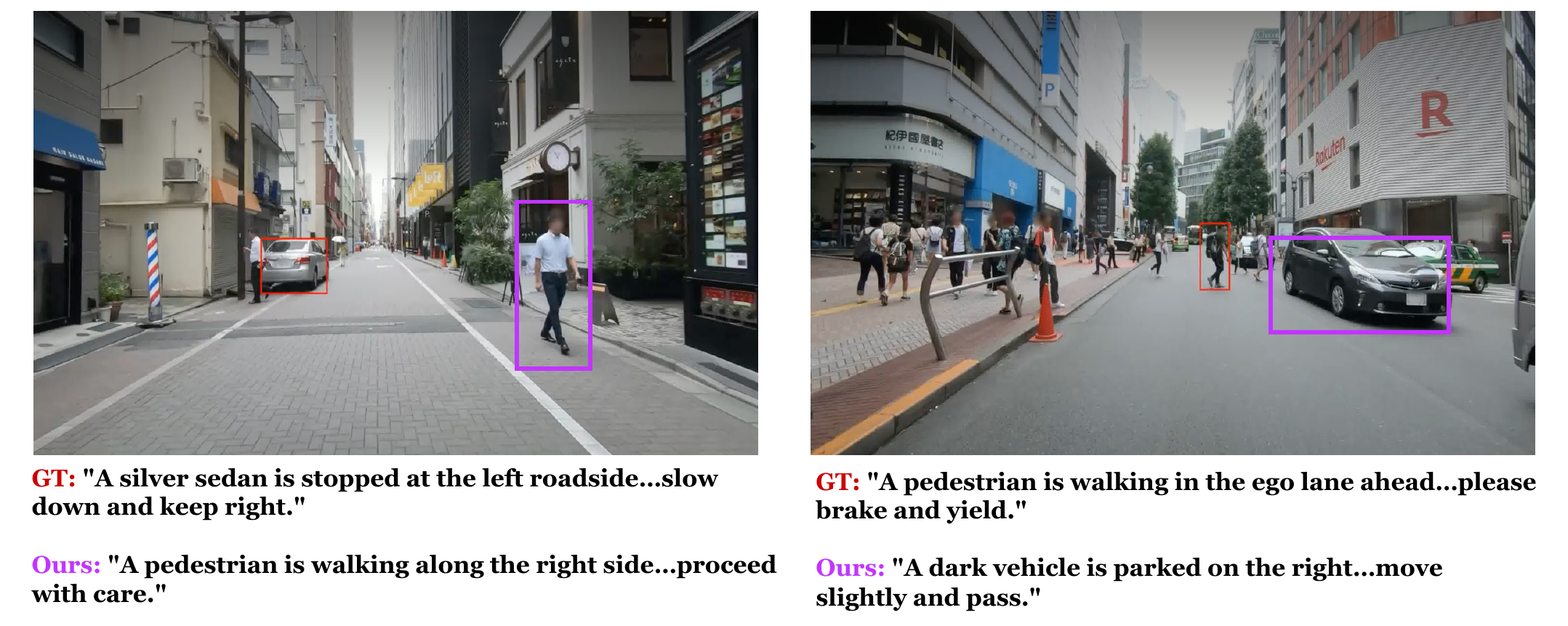}
  \caption{\textbf{Failure cases of UniDrive on the DRAMA-Reasoning val split.}
  Ground-truth risk objects are shown in red and UniDrive's predictions in purple.
  In both scenes the model grounds a plausible but non-primary hazard.
  \textbf{(Left)} The ground truth marks a sedan stopped at the left roadside,
  yet UniDrive localizes a pedestrian moving along the sidewalk, favoring a
  dynamic agent over a static obstacle.
  \textbf{(Right)} The ground truth marks a pedestrian in the ego lane, while
  UniDrive localizes a large parked vehicle and overlooks the smaller, more
  distant pedestrian on the vehicle's path. These cases indicate that, when
  multiple candidate hazards co-occur, UniDrive's risk selection can be biased
  by object size, motion, and proximity rather than by safety-critical priority.}
  \label{fig:failure}
\end{figure*}

\paragraph{Failure Case Analysis.}
A complete evaluation must also examine the conditions under which UniDrive
fails. In Figure~\ref{fig:failure}, we present two representative failure cases
from the DRAMA-Reasoning \textit{val} split. In both, the ground-truth risk
object is drawn in red and the model's prediction in purple. Notably, the two
cases share a common failure mode: rather than missing the hazard altogether,
UniDrive grounds a plausible but \emph{non-primary} object and therefore
diverges from the human-annotated primary risk.

The first case (left) involves a sedan stopped at the left roadside, which the
annotation designates as the primary risk because it constrains the drivable
space ahead of the ego-vehicle. UniDrive instead localizes a pedestrian walking
along the sidewalk and centers its explanation on this moving agent. Although
the pedestrian is a reasonable object of attention, it is not the
safety-critical target in this scenario. This behavior suggests that the model
is biased toward dynamic agents and tends to underweight static obstacles,
even when the latter more directly affect the ego-vehicle's path.

The second case (right) is, in effect, the reverse. Here the ground truth marks
a pedestrian located within the ego lane as the primary risk, whereas UniDrive
grounds a large vehicle parked at the roadside and overlooks the smaller, more
distant pedestrian. The model is drawn to a large, nearby, high-contrast object
and fails to prioritize the smaller target that lies directly on the
ego-vehicle's trajectory. This case is also a useful caveat to our earlier
results: although the high-resolution perception branch improves small-object
localization in aggregate, accurate perception alone does not guarantee correct
risk \emph{selection} when several candidate hazards co-occur.

\section{Discussion}
\label{sec:discussion}

\subsection{Grounded Risk Reasoning}

The experimental results show that UniDrive improves both language-based risk explanation and object-level grounding, but the main implication goes beyond the numerical gains themselves. The consistent improvements in captioning, mIoU, small-object localization, zero-shot transfer, and human preference suggest that safety-oriented driving understanding should not be treated as either a captioning problem or a detection problem alone. In interactive driving scenes, an explanation is only safety-relevant when it can be linked to the correct risk source, and a localized object is only interpretable when its dynamic risk relevance can be explained. This finding is consistent with the motivation of DRAMA, which emphasizes the joint role of risk localization and captioning in driving scene understanding \citep{malla2023drama}, but our results further show that such a joint formulation can be strengthened through explicit architectural alignment between temporal semantics and spatial evidence.

The comparison with image-based and video-based baselines provides a useful way to interpret this point. Image-based MLLMs can exploit fine-grained visual details in the current frame, but they have limited access to how a risk evolves over time. Video-based models, in contrast, are better positioned to capture motion and interaction cues, yet their spatial grounding may be weakened by low-resolution inputs, temporal compression, or limited visual-token budgets. This helps explain why UniDrive obtains improvements in both captioning and detection rather than only one side of the task. The high-resolution perception branch improves the model's ability to identify visually subtle or small risk objects, while the temporal reasoning branch provides dynamic context for explaining why these objects matter. As a result, the generated output becomes not only linguistically plausible, but also more visually verifiable.

The zero-shot results on NuScenes and BDD100K further indicate that the learned representation is not merely a dataset-specific mapping from DRAMA-Reasoning annotations to textual templates. The larger gains on situation reasoning and risk-object localization suggest that UniDrive captures transferable cues about object presence, motion context, and safety relevance. This is important because real-world autonomous driving systems inevitably encounter distribution shifts in weather, road layout, traffic culture, sensor viewpoint, and object appearance. Although zero-shot transfer does not prove deployment readiness, it provides evidence that grounding risk explanations in explicit visual evidence can support more robust generalization than language-only or weakly grounded reasoning.

\subsection{Role of Spatio-Temporal Fusion}

The ablation results provide insight into the mechanism behind UniDrive's performance. Removing the temporal reasoning branch reduces the quality of risk descriptions, even though the high-resolution branch can still maintain relatively strong detection performance. This suggests that high spatial fidelity alone is insufficient for interpreting dynamic driving risk. Many hazards are defined not only by what an object is, but also by how it moves, whether it is entering the ego lane, whether the ego vehicle is approaching it, and whether the surrounding context makes a cautious response necessary. In this sense, temporal reasoning contributes semantic and causal context rather than merely additional frames.

Conversely, removing the high-resolution perception branch leads to a substantial drop in detection performance, especially for small objects. This indicates that temporal context cannot compensate for missing fine-grained spatial evidence. In driving scenes, safety-critical objects may occupy only a small image region, be partially occluded, or appear at a distance. Such cases are particularly challenging for video-based MLLMs, where spatial precision may be sacrificed for temporal coverage \citep{zhang2023videollama,xu2024drivegpt4}. The results therefore support a complementary view: temporal information helps determine how a risk emerges, while high-resolution perception helps determine where the responsible object is located.

The strongest evidence for the proposed mechanism comes from the ablation of spatio-temporal fusion. Replacing gated cross-attention with naive feature concatenation causes a clear performance collapse across both captioning and grounding metrics. This shows that simply exposing the LLM to both temporal and spatial features is not enough. The model must learn how dynamic context should attend to fine-grained spatial regions. The gated cross-attention module provides this alignment mechanism by using temporal features as queries and high-resolution spatial features as keys and values. In other words, it allows the model to ask which visual regions are relevant to the evolving risk context. This mechanism explains why UniDrive can improve small-object grounding while also generating more coherent risk explanations.

This finding also refines the interpretation of recent driving-oriented MLLMs. Prior work has shown that video inputs, high-resolution images, visual question answering, and planning-oriented language supervision can each improve different aspects of autonomous driving understanding \citep{ding2023hilmd,sima2024drivelm,tian2024drivevlm,wang2023drivemlm}. Our results suggest that the next step is not simply to add more modalities, frames, or tokens, but to design risk-conditioned alignment mechanisms that connect dynamic semantics with object-level evidence. For safety-critical tasks, the key question is not only whether a model can describe a scene, but whether it can align the reason for caution with the correct visual target.

\subsection{Implications for Safety-Oriented Driving MLLMs}

In terms of model design, the results suggest that future driving MLLMs should move beyond general scene understanding and place greater emphasis on grounded risk reasoning. In general-purpose VLMs, visual grounding is often treated as a way to connect language expressions with image regions \citep{peng2023kosmos2,zhang2024groundhog,cheng2024spatialrgpt}. In autonomous driving, however, grounding carries a more direct safety meaning because it determines whether an explanation refers to the object that actually matters for the ego vehicle. Therefore, grounding should not be considered an auxiliary output attached to language generation. Instead, it should be integrated into the core reasoning process, so that language explanations, risk-object localization, and driving suggestions are mutually constrained.

This perspective also has implications for evaluation. Standard captioning metrics such as BLEU-4, METEOR, CIDEr, and SPICE are useful for measuring linguistic similarity, while IoU-based metrics are useful for measuring localization accuracy. However, neither type of metric alone can fully evaluate safety-oriented explanations. A fluent caption may refer to the wrong object, while a correct bounding box may still fail to explain why the object is risky. The combined evaluation adopted in this study, including captioning, grounding, object-scale analysis, zero-shot transfer, robustness testing, qualitative cases, and human-centered assessment, provides a more comprehensive view of model reliability. At the same time, the failure cases indicate that even this evaluation remains incomplete, because risk understanding also requires prioritizing among multiple plausible hazards. Future benchmarks should therefore consider risk ranking, trajectory relevance, causal relevance, and uncertainty, rather than evaluating only one primary bounding box or one reference caption.

The human-centered evaluation further shows why grounded explanations are important for trustworthy autonomous driving. Participants more frequently preferred UniDrive's explanations in terms of usefulness, accuracy, and trustworthiness. This result should not be interpreted merely as a preference for more fluent text. Rather, it suggests that explanations become more useful when they identify the risk object, describe the reason for concern, and provide a visually checkable basis for the recommended response. Such outputs are relevant to safety-oriented perception and post-hoc auditing, where an explanation must be inspectable rather than only plausible. In this sense, UniDrive can be viewed as an intermediate reasoning layer between perception and downstream decision-making. It does not replace planning or control, but it can provide grounded risk evidence that supports safety monitoring, human review, and failure analysis.

The robustness results further support this safety-oriented interpretation. Under night and rainy conditions, UniDrive exhibits more graceful degradation than the video-language baseline. This suggests that combining temporal context with high-resolution spatial evidence can reduce dependence on any single fragile cue. When visual details are degraded, temporal context can still provide continuity and motion information; when temporal cues are ambiguous, high-resolution perception can still preserve object-level evidence. Such complementarity is particularly important for safety-relevant scenarios, where rare, adverse, or long-tail conditions often determine system reliability.

\subsection{Limitations and Future Studies}

Despite these promising results, several limitations remain. First, the failure cases reveal that UniDrive can still select a plausible but non-primary risk object when multiple candidate hazards co-occur. This is not a simple object detection failure. In both representative cases, the model grounds an object that is visually reasonable, but not the one annotated as most safety-critical. This suggests a risk-prioritization problem. The model may be biased toward dynamic agents such as pedestrians, or toward large, nearby, and visually salient objects, even when another object has stronger relevance to the ego vehicle's trajectory. Future work should therefore move beyond single-object grounding and introduce risk hierarchy, multi-risk ranking, and trajectory-aware relevance estimation. Explicit risk indicators such as time-to-collision, post-encroachment time, collision probability, reachable sets, or driving-corridor constraints could help distinguish visually salient objects from genuinely safety-critical ones.

Second, the current DRAMA-Reasoning setting is built on primary risk-object annotations and enriched textual descriptions. This design enables joint evaluation of reasoning and grounding, but it also simplifies the complexity of real driving scenes. In practice, multiple objects may simultaneously contribute to risk at different levels, and different human annotators may disagree about which object should be considered primary. Future datasets should therefore include multiple risk candidates, risk severity levels, causal relations, and inter-annotator agreement. Such annotations would make it possible to evaluate whether a model understands the structure of risk rather than only matching one selected target.

Third, UniDrive is primarily based on visual-language reasoning and does not explicitly incorporate all information used by a full autonomous driving stack. Although the model can infer many risk cues from video and high-resolution images, it does not directly model ego-motion, vehicle dynamics, traffic rules, HD maps, LiDAR/radar measurements, or downstream planning feasibility. This limits its ability to determine which object will affect the ego vehicle's future path under different maneuvers. Future studies should integrate grounded risk reasoning with BEV representations, map priors, trajectory prediction, and planning modules. Such integration would allow the model to reason not only about what is visible, but also about what is reachable, avoidable, and operationally relevant.

Finally, deployment efficiency and reliability require further investigation. UniDrive achieves a reasonable balance between performance and computational cost, but the reported inference speed on an A100 GPU should not be interpreted as sufficient evidence for real-time in-vehicle deployment. Practical systems require lower latency, stronger reliability guarantees, and extensive validation under closed-loop conditions. Future work should explore model compression, knowledge distillation, token pruning, and edge-oriented inference. More importantly, the value of grounded risk explanations should be tested in closed-loop simulation, human-in-the-loop experiments, and real-world replay studies to examine whether they improve downstream safety monitoring, decision support, and failure diagnosis.

\section{Conclusion}
\label{sec:conclusion}

This paper presented UniDrive, a unified visual-language and grounding framework for interpretable risk understanding in autonomous driving. Motivated by the persistent trade-off between temporal understanding and spatial precision in existing driving-oriented MLLMs, UniDrive integrates a temporal reasoning branch, a high-resolution perception branch, and a gated cross-attention fusion module to align dynamic scene context with fine-grained spatial evidence. Based on this fused representation, the model jointly generates natural-language risk descriptions and grounded bounding-box outputs for risk-critical objects.

Experiments on the DRAMA-Reasoning benchmark demonstrate that UniDrive achieves strong performance in both driving scene captioning and risk-object localization. Compared with representative image-based and video-based baselines, UniDrive provides more accurate grounded explanations, stronger small-object localization, and better overall reasoning-grounding consistency. Additional evaluations further show its zero-shot generalization to unseen driving datasets, improved robustness under adverse conditions, and higher human-rated usefulness, accuracy, and trustworthiness. Ablation studies confirm that temporal reasoning, high-resolution perception, and explicit spatio-temporal fusion are all essential to the final performance.

Overall, the findings show that safety-oriented driving MLLMs should not be evaluated only by their ability to generate plausible language responses or detect objects in isolation. Instead, they should be able to explain why a situation is risky and ground that explanation in the correct visual evidence. By linking temporal semantics, high-resolution perception, and object-level grounding within a unified generative framework, UniDrive provides a more transparent and reliable foundation for interpretable autonomous driving intelligence.

\section*{Statement}

\begin{enumerate}[label=\arabic*.]
\item \textbf{Conflict of Interest} \\
All authors declare no financial or non-financial competing interests.
\item \textbf{Data Availability} \\
Data will be made available on reasonable request.
\item \textbf{Contributions} \\
Xiaowei Gao: Conceptualization, Methodology, Data curation, Software, Formal analysis, and Writing - original draft. Pengxiang Li: Conceptualization, Methodology, Data curation, Software, Formal analysis, and Writing - original draft. Yitai Cheng: Visualization, and Writing - review \& editing. Ruihan Xu: Data curation, and Writing - review \& editing. James Haworth: Writing - review \& editing, and Validation. Stephen Law: Writing - review \& editing, and Validation. Yun Ye: Conceptualization, Methodology, Resources, Writing - review \& editing, Project administration, and Supervision. All authors read and approved the final manuscript.
\item \textbf{Declaration of the use of AI} \\
During the preparation of this work the authors used ChatGPT in
order to improve the readability and language of the manuscript. After
using this tool/service, the authors reviewed and edited the content as
needed and take full responsibility for the content of the publication.

\end{enumerate}

\printcredits

\bibliographystyle{cas-model2-names}

\bibliography{cas-refs}



\end{document}